\documentclass{article}

\usepackage{graphicx} 
\usepackage{subfigure}

\usepackage{natbib}

\usepackage{algorithm}
\usepackage{algorithmic}

\usepackage{amssymb}
\usepackage{mathtools}
\usepackage{xfrac}

\usepackage{hyperref}
\usepackage[all]{hypcap}	

\usepackage[accepted]{icml2012}

\icmltitlerunning{Predicting Drug Interactions and Mutagenicity with Ensemble Classifiers on Subgraphs of Molecules}

\begin{document}

\twocolumn[
\icmltitle{Predicting Drug Interactions and Mutagenicity with Ensemble Classifiers on Subgraphs of Molecules}

\small
{\fontsize{8.5}{9}\icmlauthor{Andrew~Schaumberg\textsuperscript{1}*, Angela~Yu\textsuperscript{1}, Tatsuhiro~Koshi\textsuperscript{2}, Xiaochan~Zong\textsuperscript{3}, and Santoshkalyan~Rayadhurgam\textsuperscript{4}}{}}
\begin{center}
\icmladdress{\textsuperscript{1}\textit{Cornell University and the Tri-Institutional Training Program in Computational Biology and Medicine} \\ \textsuperscript{2}\textit{Weill Institute for Cell and Molecular Biology and Department of Molecular Biology and Genetics, Cornell University} \\ \textsuperscript{3}\textit{School of Electrical and Computer Engineering, Cornell University} \\ \textsuperscript{4}\textit{Department of Materials Science and Engineering, Cornell University} \\ Cornell University, Day Hall, Ithaca, NY 14853 USA \\ (Dated: Dec 11, 2014)}
\end{center}
\normalsize

\icmlkeywords{machine learning, graph kernel, molecular interaction}

\vskip 0.3in
]

\begin{abstract}
In this study, we intend to solve a mutual information problem in interacting molecules of any type, such as proteins, nucleic acids, and small molecules. Using machine learning techniques, we accurately predict pairwise interactions, which can be of medical and biological importance. Graphs are are useful in this problem for their generality to all types of molecules, due to the inherent association of atoms through atomic bonds. Subgraphs can represent different molecular domains. These domains can be biologically significant as most molecules only have portions that are of functional significance and can interact with other domains. Thus, we use subgraphs as features in different machine learning algorithms to predict if two drugs interact and predict potential single molecule effects.
\end{abstract}

\section*{Introduction}
In drug discovery, each new molecular entity costs on average \$1.8 billion \cite{Paul10}. Finding drug candidates from over 50 million small molecules available in open access databases ~\cite{Roy10}, such as PubChem ~\cite{Bolton08} and ChEMBL ~\cite{Gaulton12}, have the potential to save both lives and money.

Over 90,000 proteins ~\cite{Porta13} and 40,000 small molecule metabolites ~\cite{Wishart13} naturally occur in human cells.  Function is often characterized by interactions, e.g. KEGG ~\cite{Kanehisa00} defines functional biological pathways in terms of pairwise molecular interactions, while DrugBank ~\cite{Law14} defines drug-drug interactions that influence dosage and functional drug-target interactions.  Machine learning efficiently finds promising drug candidates, reducing drug discovery cost by only taking drugs to in vitro tests that are predicted to selectively interact with the molecule of interest ~\cite{Fukunishi09} \cite{Khandelwal08}.

\section*{Problem Definition and Methods}
In our first aim, we benchmark our methods on a mutagenicity prediction dataset.  In our second aim, we predict the interactions of interest.

\subsection*{Task Definition}
\begin{description}
\item[Aim 1] Predict mutagenicity based on subgraphs.
Mutagenicity is the ability of a compound to change genetic materials. If a drug is mutagenic, it cannot be safely consumed by a patient, even with predicted drug-target interaction.
\item[Aim 2] Predict drug interactions based on their subgraphs.  We consider both small molecules and proteins for interactions.  An interaction occurs if dosage must be adjusted, due to both drugs being present in a patient ~\cite{Law14}.
\end{description}

\subsection*{Algorithms and Methods}

Features consist of subgraph counts derived from the dataset. Subgraphs to be included in the feature set depended on parameter choices of height and distance. Height describes the max distance between any two nodes in the subgraph. Distance is the max number of nodes away two subgraphs are from each other in the full graph.

\begin{figure}[]
\begin{center}
\begin{small}
\begin{sc}
\includegraphics[width=0.5\textwidth]{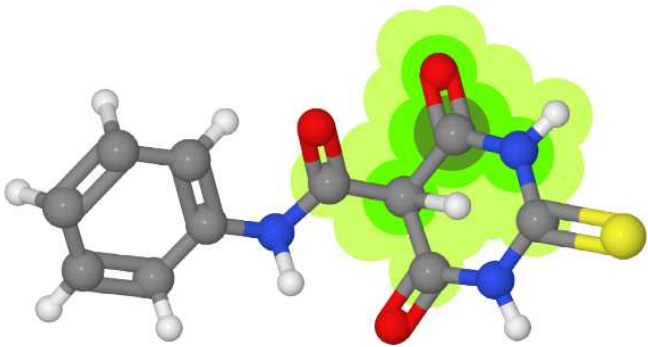}
\caption{Subgraph heights 0, 1, and 2 in dark, light, and pale green regions, respectively}
\end{sc}
\end{small}
\end{center}
\label{fig:feat_heights}
\end{figure}

Rather than use only subgraphs at a single height, we subgraphs for multiple heights up to a maximum height (Figure ~\ref{fig:feat_heights}).  We let subgraph \textit{s} at height \textit{h} among atoms \textit{A} be defined as:
\begin{equation}
s(h,a \in A,A) = \forall b \in A, d(a, b) \leq h
\end{equation}
where \textit{d} is shortest path distance.  Let \textit{c} be the count of occurrences of a subgraph \textit{s} in a molecule \textit{M} having atoms \textit{A}:
\begin{equation}
c(h,A) = \sum_{a \in A}^{} I(s(h, a, A))
\end{equation}
where \textit{I} is the indicator function.  We let the feature set \textit{F} of all counts \textit{c} at desired heights \textit{H} fully describe the molecule \textit{M} having atoms \textit{A}:
\begin{equation}
F(H,A \in M) = \forall h \in H, \{ c_1(h,A), ..., c_{n=|F|}(h,A) \}
\end{equation}

For mutagenicity prediction, we considered two kinds of feature sets:
\begin{enumerate}
\item Similar to prior published methods ~\cite{Grave10a}, we considered a pair of subgraphs separated by a distance as a feature.
\item More simply, we considered subgraphs at desired heights as features, rather than subgraph pairs.
\end{enumerate}

For drug interaction prediction, our method reduces computational cost in two ways:
\begin{enumerate}
\item Consider only a subset of heights up to a maximum height, not all such heights, and
\item do not consider pairs of subgraphs separated by a distance as features.
\end{enumerate}

We also implemented the two kernels from Costa and Grave (2010) and Faulon et al. (2008) and generated kernelized and un-kernelized data ~\cite{Grave10a} ~\cite{Faulon08}. We were unsure as to the possibility of  implementing a kernel trick into all of the ensemble classifiers we chose to use, so we had to generate kernelized input data to use in our ensemble classifiers.  Thus used three types of datasets:
\begin{enumerate}
\item Subgraphs or subgraph pair data.
\item Data after cosine similarity kernel applied, like Faulon 2008.
\item Data after NSPDK kernel applied, like Costa and Grave 2010.
\end{enumerate}

We tested convolutional neural network (CNN), random forest (RF), multilayer perceptron (MLP), and support vector machine (SVM) algorithms.  Random forest performed best.

\begin{table*}[th]
\caption{Machine learning algorithm performance: area under ROC and validation set accuracy.  Each algorithm's performance is noted for the top one or two height/distance formats parsed from the Bursi dataset. RF metrics listed as mean $\pm$ stdev. MLP and CNN metrics listed as (min,mean$\pm$stdev,max). CNN processes each data row as three partitions, each partition being one third of the original row, the first third being subgraphs with the fewest atoms, the last third being subgraphs with the most atoms.  This CNN connects the first partition to the first hidden node, the first and second partitions to the second hidden node, the second and third partitions to the third hidden node, and the third partition to the fourth hidden node. For MLP, 3$\rightarrow$1 topology means there are three hidden nodes and one output node.  For RF, 100 trees were in the forest, where each tree was allowed to grow out fully.  Kernels were not used for any algorithm reported in this table.}
\label{sample-table2}
\vskip 0.15in
\begin{small}
\begin{sc}
\begin{center}
\begin{tabular}{lcccccr}
\hline
\abovespace\belowspace
Algo & Topology & Height/Distance & AUROC & Validation Acc & Trials \\
\hline
\abovespace
RF & data$\rightarrow$100tr & H=1 & 0.89739$\pm$0.024 & 0.82639$\pm$0.011 & 100 \\
RF & data$\rightarrow$100tr & H=[0-3] D=[0-5] & 0.89486$\pm$0.007 & 0.82022$\pm$0.009 & 100 \\
CNN & 3(\sfrac{data}{3})$\rightarrow$4$\rightarrow$1 & H=[0-3] D=[0-5] & - & (0.78701,0.79085$\pm$0.004,0.79609) & 4 \\
MLP & data$\rightarrow$3$\rightarrow$1 & H=2 & - & (0.60545,0.76383$\pm$0.044,0.81564) & 40 \\
MLP & data$\rightarrow$3$\rightarrow$1 & H=[0-3] D=[0-5] & - & (0.55307,0.72895$\pm$0.108,0.82332) & 7 \\
\hline
\end{tabular}
\end{center}
\end{sc}
\end{small}
\vskip -0.1in
\end{table*}

\subsection*{Random Forest}
Random forests were generated using Scikit-learn’s RandomForestClassifier ~\cite{scikit-learn}. Random forest subsamples the training data and feature set to build each decision tree in the forest. Each decision tree was allowed to grow fully and both entropy and gini impurity were tried as splitting criteria. 100 decision trees were used to make the random forest. This particular implementation averages the probabilistic prediction of all decision trees in the forest instead of letting each decision tree vote for a classification. 

\subsection*{Support Vector Machine}
We used SVMlight on all dataset types with all combinations of settings:
\begin{enumerate}
\item Linear hyperplane (\texttt{svm\_learn -t 0}) or radial basis function kernel (\texttt{svm\_learn -t 2}) classifier.
\item Negative example ratio correction enabled or disabled (\texttt{svm\_learn -j 0.8}).  In the 
Bursi data, there are 0.8 negative examples for every positive example.
\item Sum of slacks, C, of 1, 10, and 100 (\texttt{svm\_learn -c 1}).
\end{enumerate}

\subsection*{Multilayer Perceptron}
We implemented a multilayer perceptron ~\cite{MachineLearning}:
\begin{enumerate}
\item One output node and three hidden nodes.
\item Weight updates through backpropagation and momentum.
\item Voted and unvoted perceptrons.
\item Annealed learning rate.
\item Convergence when validation error did not improve after one full iteration through data.
\item For online learning, not batch learning.
\end{enumerate}

\subsection*{Convolutional Neural Network}
We implemented a convolutional neural network ~\cite{LeCun98}:
\begin{enumerate}
\item One output node and four hidden nodes.
\item Data split into thirds by subgraph mass, e.g. when sorting a row’s X subgraphs by lowest to highest mass, the \sfrac{X}{3} subgraphs are grouped as the lightest third, the next \sfrac{X}{3} subgraphs are grouped into the middle third, and the next \sfrac{X}{3} subgraphs are grouped into the heaviest third.
\item Other settings identical to MLP.
\end{enumerate}

\section*{Experimental Evaluation}
\subsection*{Methodology}
To allow for direct comparison to prior methods ~\cite{Grave10a} ~\cite{Faulon08}, we optimized for area under the receiver operator curve (AUROC) instead of accuracy. AUROC is a measure of the effect of changing the discrimination threshold of a binary classifier. It is calculated by plotting the true positive rate against the false positive rate over many different threshold values, and then calculate the area under said curve. AUROC and accuracy are different because accuracy can be thought of as a measurement at one threshold value. For completeness, we report training and validation accuracy. We used both 10-fold cross-validation and 100 iterations of randomly splitting the dataset such that \sfrac{2}{3} makes up a training set and \sfrac{1}{3} makes up a validation set. We also compare different methods using two-sample un-pooled t-tests with a significance threshold (alpha) of 0.05. 

\subsubsection*{Mutagenicity Prediction}
We computed features at a variety of feature heights (1-12) and distances (0-20). Algorithms we tested include RF, MLP, CNN, and SVM. Statistical significances were determined using un-pooled two-sample t-tests.

We wrote a Bursi data parser that processes the data in two ways: (1) decomposing a molecule into subgraphs \cite{Faulon08}, and (2) decomposing a molecule into subgraph pairs separated at a distance \cite{Grave10a}. Though in prior work ~\cite{Grave10a}, we forwent PTC and NTP HIV analyses for two reasons: (1) their categorical classifications are more complex than the binary classifications in Bursi, and (2) our motivation to focus on interaction prediction for the public good.

\begin{figure}[t]
\begin{center}
\begin{small}
\begin{sc}
\includegraphics[width=0.5\textwidth]{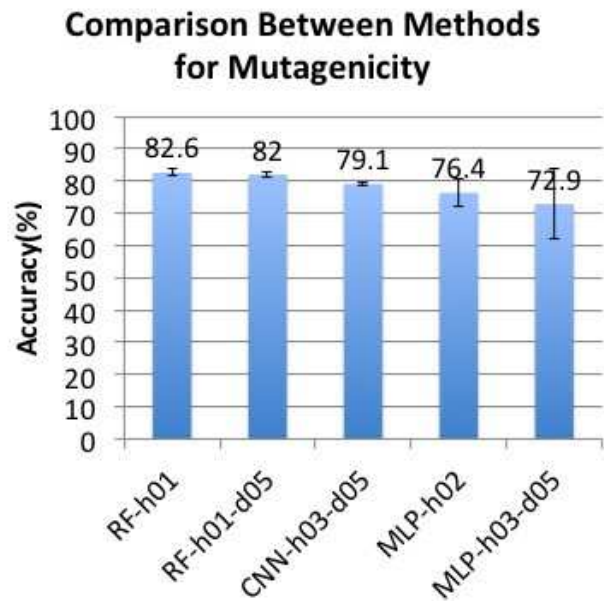}
\caption{Comparison Between Methods for Mutagenicity}
\end{sc}
\end{small}
\end{center}
\label{fig:muta_meths_ours}
\end{figure}

\subsubsection*{Interaction Prediction}
Means and standard deviations of AUROC and accuracies were established using 10-fold cross-validation. 

We wrote a DrugBank data parser that decomposes a molecule into subgraphs.  For drugs or targets listed in DrugBank where SMILES or protein sequence was not provided, the parser fetches the missing information from PubChem or UniProt, respectively.  PubChem results where the chemical name does not match the DrugBank-provided name are rejected.  We took two samples from DrugBank for evaluation: (1) 37 positive training examples and 39 negative training examples, and (2) 131 positive training examples and 140 negative training examples.

\subsection*{Results}

\begin{figure}[h]
\begin{center}
\begin{small}
\begin{sc}
\includegraphics[width=0.5\textwidth]{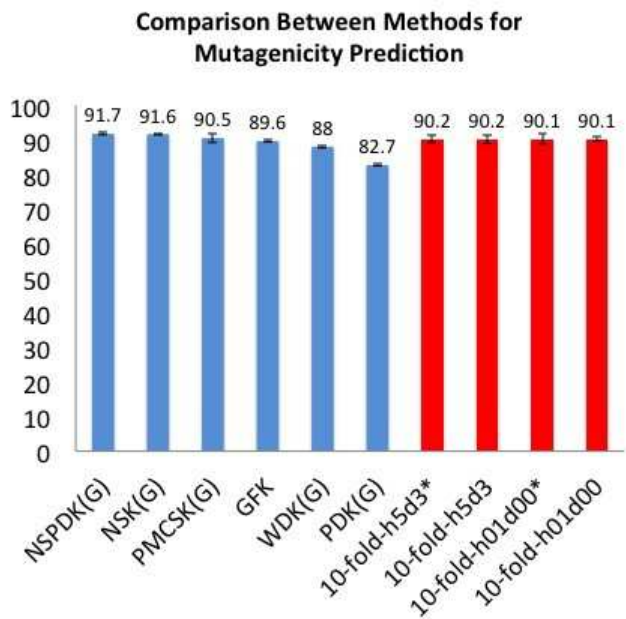}
\caption{Comparison Between Methods for Mutagenicity Prediction}
\end{sc}
\end{small}
\end{center}
\label{fig:muta_meths_full}
\end{figure}

\begin{figure}[ht]
\begin{center}
\begin{small}
\begin{sc}
\includegraphics[width=0.5\textwidth]{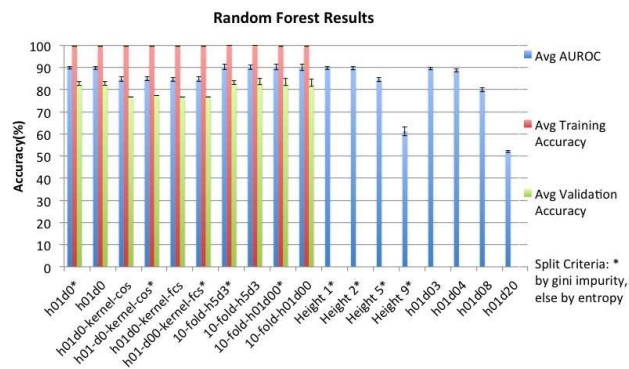}
\caption{Random Forest Results}
\end{sc}
\end{small}
\end{center}
\label{fig:rf}
\end{figure}

\begin{figure}[ht]
\begin{center}
\begin{small}
\begin{sc}
\includegraphics[width=0.5\textwidth]{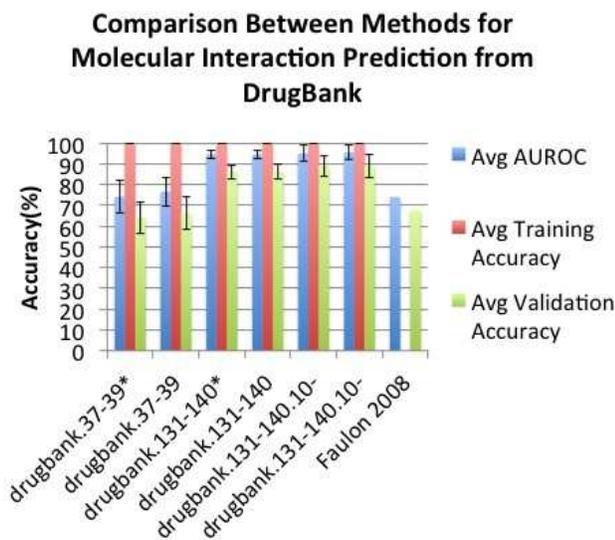}
\caption{Comparison Between Methods for Molecular Interaction Prediction from DrugBank}
\end{sc}
\end{small}
\end{center}
\label{fig:drug_meths_full}
\end{figure}

Among all the algorithms that we tried, random forest had the best result out of the ensemble classifiers (Table 1) (Figure~\ref{fig:muta_meths_ours}). Random forest did significantly better than the other ensemble classifiers (from two-sample unpooled t-tests). A more detailed analysis of the random forest method showed that using height one gave us the best results (Figure~\ref{fig:rf}). Also the two different splitting criteria did not produce any significant changes (from two-sample unpooled t-tests). We compared random forest and previously published methods for mutagenicity prediction, and found that random forest on un-kernelized height one and distance zero data did as well as or better than four previously published kernel-based methods (Figure~\ref{fig:muta_meths_full}). However random forest is not significantly better than any of the previously published methods (from two-sample unpooled t-tests). We also compared our random forest method against previously published methods for molecular interaction prediction from DrugBank data. The drug-target interactions with random forest had 94.5\% AUROC and 86.1\% validation accuracy (Figure~\ref{fig:drug_meths_full}) while a previously published kernel based method got 74\% AUROC and 67.5\% accuracy when they optimized for AUROC ~\cite{Faulon08}. We cannot establish if our result is statistically significant because the other method did not report a standard deviation. However we believe ~20\% difference probably is statistically significant. 

\subsection*{Discussion}

\textit{Random forest on un-kernelized data does as well as or better than four previously published kernel-based methods in prediction mutagenicity.} 
We hypothesize that this may be due to small local rules (such as ``domain A with domain B is always mutagenic'') in chemistry that are not well approximated in the full dataset. Random forest uses subsets of training data and subsets of features to build the decision trees, which is more likely to capture these types of rules. The random forest implementation we used allows each decision tree to grow out fully, so possible features that can be used as splitting criteria depends on the subset of features used to generate each tree. This allows each tree to try to approximate potentially different small biological/chemistry rules. Height 1 and distance 0 produced the best results, which may seem counterintuitive at first because the subgraphs themselves do not contain much information on the molecule’s topology. However, structure of individual trees may imply neighboring structures. Molecules are 3-dimensional structures, but we parsed and used their 2-dimensional structure so neighboring subgraphs in 3-dimensional space may not be captured with subgraph representation alone. Every splitting criteria in each tree of the random forest is a feature representing the count of a small subgraph. Therefore, series of splits might represent a local structure found in the whole molecule. Another seemingly odd result we encountered is that using kernelized data did not help the AUROC or accuracies of our ensemble classifiers. The kernels we used were designed for use in SVMs and may not translate well into ensemble classifiers. 

\textit{Drug-target interactions with random forest gets 94.5\% AUROC and 86.1\% validation accuracy while a previously published kernel based method got 74\% AUROC and 67.5\% accuracy.} 
The other method used slightly different data that we tried to replicate as best as possible. We hypothesize that this may be due to small local rules (such as ``domain A interacts with domain B’’) in chemistry that are not well approximated in the full dataset. Reasons outlined in the above as to why we believe random forest did better than previously published methods and why kernelizing the data did not help also applies here to molecular interaction prediction. 

\section*{Related Work}

Costa and Grave (2010) did not predict drug-target interactions, but used all feature heights up to a maximum height and all distances separating subgraph pairs up to a maximum distance ~\cite{Grave10a}. The authors achieved 91.7\% AUROC for the graphs not augmented with functional groups from organic chemistry, improving to 92.0\% AUROC with augmentation.  Parsing all features for all small and large molecules in DrugBank was not computationally feasible in terms of time or memory.

Faulon et al. (2008) predicted drug-target interactions using support vector machines and a cosine similarity kernel, achieving 74\% AUROC and 67.5\% accuracy when optimizing for AUROC ~\cite{Faulon08}.  The authors used features at a single height for small molecules and a single height for proteins, rather than multiple heights.

Yu et al. (2012) also predicted drug-target interactions using random forest classifiers and achieved an average 89.65\% AUROC ~\cite{Yu12}. However, the feature set they used consisted of chemical and protein descriptors instead of subgraphs as in our project. Their dataset also is different than ours, but it is interesting to note that random forest has been applied to similar problem setups and has done well. 

\section*{Future Work}
\begin{enumerate}
\item We kernelized the data without considering interaction terms between features, so the next step is to kernelize the expanded feature set as well as trying to use the un-kernelized expanded feature set.
\item Try reducing the feature set based on results from a random forest before using other ensemble classifiers and kernels. Random forests can be used to assign variable importance to features. By using a reduced feature set, we might be able to more accurately capture small rules and potentially increase the accuracy of all the methods we tried in this project. 
\item Use augmented graphs with annotated nodes and edges, using annotations such as domain family. Costa and Grave (2010) found a small improvement to their AUROC when using an augmented graph ~\cite{Grave10a}. 
\item Take a larger sample from DrugBank.  We are currently parsing 20,000+ positive training examples.
\item Continue developing ensemble classifier-based methods with kernels.
\end{enumerate}

\section*{Conclusion}

Random forest performed well in mutagenicity prediction and molecular interaction prediction compared to previously published methods. We show that random forest performs better than other ensemble classifiers in mutagenicity prediction and that parameter settings of height one, distance 0, and un-kernelized data produced the best results. Our random forest implementation did not do significantly better than previously published methods we found, but in terms of mean AUROC $\pm$ one standard deviation, we did as well or better than 4 out of 6 other methods ~\cite{Grave10a}. Our random forest implementation for molecular interaction prediction did around 20\% better in terms of AUROC and accuracy than a previously published method ~\cite{Faulon08}. 

\section*{Acknowledgements}

Support was provided by the Tri-Institutional Training Program in Computational Biology and Medicine (via NIH training grant T32GM083937.
The authors thank Professor Thorsten Joachims for teaching the class, and Adith Swaminathan for reviewing the project.

\small
\bibliography{example_paper}

\begin{thebibliography}{16}
\providecommand{\natexlab}[1]{#1}
\providecommand{\url}[1]{\texttt{#1}}
\expandafter\ifx\csname urlstyle\endcsname\relax
  \providecommand{\doi}[1]{doi: #1}\else
  \providecommand{\doi}{doi: \begingroup \urlstyle{rm}\Url}\fi

\bibitem[Bolton et~al.(2008)Bolton, Wang, Thiessen, and Bryant]{Bolton08}
Bolton, E., Wang, Y., Thiessen, P.~A., and Bryant, S.~H.
\newblock {P}ub{C}hem: {I}ntegrated platform of small molecules and biological
  activities.
\newblock \emph{Annual Reports in Computational Chemistry}, 4\penalty0
  (12):\penalty0 1--27, 2008.
\newblock \htmladdnormallink{[ACS
  manuscript]}{http://oldwww.acscomp.org/Publications/ARCC/volume4/chapter12.html},\htmladdnormallink{[author
  manuscript]}{ftp://ftp.ncbi.nlm.nih.gov/pubchem/publications/ARCC_PubChem_Integrated_Platform.pdf}.

\bibitem[Costa \& Grave(2010)Costa and Grave]{Grave10a}
Costa, F. and Grave, K.~D.
\newblock Fast neighborhood subgraph pairwise distance kernel.
\newblock In \emph{Proceedings of the 27th International Conference on Machine
  Learning (ICML 2010)}, Haifa, Israel, 2010.
\newblock \htmladdnormallink{[ICML
  manuscript]}{http://www.icml2010.org/papers/347.pdf}.

\bibitem[Faulon et~al.(2008)Faulon, Misra, Martin, Sale, and Sapra]{Faulon08}
Faulon, J.~L., Misra, M., Martin, S., Sale, K., and Sapra, R.
\newblock Genome scale enzyme-metabolite and drug-target interaction
  predictions using the signature molecular descriptor.
\newblock \emph{Bioinformatics}, 2\penalty0 (50):\penalty0 225--233, 2008.
\newblock
  \htmladdnormallink{PMID:18037612}{https://www.ncbi.nlm.nih.gov/pubmed/18037612}.

\bibitem[Fukunishi(2009)]{Fukunishi09}
Fukunishi, Y.
\newblock Structure-based drug screening and ligand-based drug screening with
  machine learning.
\newblock \emph{Combinatorial chemistry \& High Throughput Screening},
  12\penalty0 (4):\penalty0 397--408, 2009.
\newblock
  \htmladdnormallink{PMID:19442067}{http://www.ncbi.nlm.nih.gov/pubmed/19442067}.

\bibitem[Gaulton et~al.(2012)Gaulton, Bellis, Bento, Chambers, Davies, Hersey,
  Light, McGlinchey, Michalovich, Al-Lazikani, and Overington]{Gaulton12}
Gaulton, A., Bellis, L.~J., Bento, A.~P., Chambers, J., Davies, M., Hersey, A.,
  Light, Y., McGlinchey, S., Michalovich, D., Al-Lazikani, B., and Overington,
  J.~P.
\newblock {ChEMBL}: a large-scale bioactivity database for drug discovery.
\newblock \emph{Nucleic Acids Research}, 40\penalty0 (Database issue):\penalty0
  1100--1107, 2012.
\newblock
  \htmladdnormallink{PMID:21948594}{https://www.ncbi.nlm.nih.gov/pubmed/21948594}.

\bibitem[Gonz\`{a}lez-Porta et~al.(2013)Gonz\`{a}lez-Porta, Frankish, Rung,
  Harrow, and Brazma]{Porta13}
Gonz\`{a}lez-Porta, M., Frankish, A., Rung, J., Harrow, J., and Brazma, A.
\newblock {T}ranscriptome analysis of human tissues and cell lines reveals one
  dominant transcript per gene.
\newblock \emph{Genome Biology}, 14\penalty0 (R70):\penalty0 1--11, 2013.
\newblock
  \htmladdnormallink{PMID:23815980}{http://www.ncbi.nlm.nih.gov/pubmed/23815980}.

\bibitem[Kanehisa \& Goto(2000)Kanehisa and Goto]{Kanehisa00}
Kanehisa, M. and Goto, S.
\newblock {KEGG} -- kyoto encyclopedia of genes and genomes.
\newblock \emph{Nucleic Acids Research}, 28\penalty0 (1):\penalty0 27--30,
  2000.
\newblock
  \htmladdnormallink{PMID:10592173}{http://www.ncbi.nlm.nih.gov/pubmed/10592173}.

\bibitem[Khandelwal et~al.(2008)Khandelwal, Krasowski, Reschly, Sinz, Swaan,
  and Ekins]{Khandelwal08}
Khandelwal, A., Krasowski, M.D., Reschly, E.J., Sinz, M.W., Swaan, P.W., and
  Ekins, S.
\newblock Machine learning methods and docking for predicting human pregnane x
  receptor activation.
\newblock \emph{Chemical Research in Toxicology}, 21\penalty0 (7):\penalty0
  1457--1467, 2008.
\newblock
  \htmladdnormallink{PMID:18547065}{http://www.ncbi.nlm.nih.gov/pubmed/18547065}.

\bibitem[Law et~al.(2014)Law, Knox, Djoumbou, Jewison, Guo, Liu, Maxiejewski,
  Arndt, Wilson, Neveu, Tang, Gabriel, Ly, Adamjee, Dame, Han, Zhou, and
  Wishart]{Law14}
Law, V., Knox, C., Djoumbou, Y., Jewison, T., Guo, A.~C., Liu, Y., Maxiejewski,
  A., Arndt, D., Wilson, M., Neveu, V., Tang, A., Gabriel, G., Ly, C., Adamjee,
  S., Dame, Z.~T., Han, B., Zhou, Y., and Wishart, D.~S.
\newblock Drugbank 4.0: shedding new light on drug metabolism.
\newblock \emph{Nucleic Acids Research}, 42\penalty0 (D):\penalty0 1091--1097,
  2014.
\newblock
  \htmladdnormallink{DOI:10.1093/nar/gkt1068}{http://dx.doi.org/10.1093/nar/gkt1068},\htmladdnormallink{PMID:24203711}{http://www.ncbi.nlm.nih.gov/pubmed/24203711}.

\bibitem[LeCun et~al.(1998)LeCun, Bottou, Bengio, and Haffner]{LeCun98}
LeCun, Y., Bottou, L., Bengio, Y., and Haffner, P.
\newblock Gradient-based learning applied to document recognition.
\newblock \emph{Proceedings of the IEEE}, 86\penalty0 (11):\penalty0
  2278--2324, 1998.
\newblock
  \htmladdnormallink{DOI:10.1109/5.726791}{http://dx.doi.org/10.1109/5.726791},\htmladdnormallink{[author
  manuscript]}{http://yann.lecun.com/exdb/publis/pdf/lecun-01a.pdf}.

\bibitem[Mitchell(1997)]{MachineLearning}
Mitchell, Tom.~M.
\newblock \emph{Machine Learning}.
\newblock McGraw-Hill, 1997.
\newblock ISBN 0070428077.

\bibitem[Paul et~al.(2010)Paul, Mytelka, Dunwiddie, Persinger, Munos, Lindborg,
  and Schacht]{Paul10}
Paul, S.~M., Mytelka, D.~S., Dunwiddie, C.~T., Persinger, C.~C., Munos, B.~H.,
  Lindborg, S.~R., and Schacht, A.~L.
\newblock {H}ow to improve {R}\&{D} productivity: the pharmaceutical industry's
  grand challenge.
\newblock \emph{Nature Reviews Drug Discovery}, 9\penalty0 (9):\penalty0
  203--214, 2010.
\newblock
  \htmladdnormallink{PMID:20168317}{http://www.ncbi.nlm.nih.gov/pubmed/20168317}.

\bibitem[Pedregosa et~al.(2011)Pedregosa, Varoquaux, Gramfort, Michel, Thirion,
  Grisel, Blondel, Prettenhofer, Weiss, Dubourg, Vanderplas, Passos,
  Cournapeau, Brucher, Perrot, and Duchesnay]{scikit-learn}
Pedregosa, F., Varoquaux, G., Gramfort, A., Michel, V., Thirion, B., Grisel,
  O., Blondel, M., Prettenhofer, P., Weiss, R., Dubourg, V., Vanderplas, J.,
  Passos, A., Cournapeau, D., Brucher, M., Perrot, M., and Duchesnay, E.
\newblock Scikit-learn: {M}achine {L}earning in {P}ython.
\newblock \emph{Journal of Machine Learning Research}, 12:\penalty0 2825--2830,
  2011.
\newblock \htmladdnormallink{[JMLR
  manuscript]}{http://jmlr.csail.mit.edu/papers/v12/pedregosa11a.html}.

\bibitem[Roy et~al.(2010)Roy, McDonald, Sittampalam, and Chaguturu]{Roy10}
Roy, A., McDonald, P.~R., Sittampalam, S., and Chaguturu, R.
\newblock Open access high throughput drug discovery in the public domain: a
  {M}ount {E}verest in the making.
\newblock \emph{Current Pharmaceutical Biotechnology}, 7\penalty0
  (11):\penalty0 764--778, 2010.
\newblock
  \htmladdnormallink{PMID:20809896}{http://www.ncbi.nlm.nih.gov/pubmed/20809896}.

\bibitem[Wishart et~al.(2013)Wishart, Jewison, Guo, Wilson, Knox, Liu,
  Djoumbou, Mandal, Aziat, Dong, Bouatra, Sinelnikov, Arndt, Xia, Liu, Yallou,
  Bjorndahl, Perez-Pineiro, Eisner, Allen, Neveu, Greiner, and
  Scalbert]{Wishart13}
Wishart, D.~S., Jewison, T., Guo, A.~C., Wilson, M., Knox, C., Liu, Y.,
  Djoumbou, Y., Mandal, R., Aziat, F., Dong, E., Bouatra, S., Sinelnikov, I.,
  Arndt, D., Xia, J., Liu, P., Yallou, F., Bjorndahl, T., Perez-Pineiro, R.,
  Eisner, R., Allen, F., Neveu, V., Greiner, R., and Scalbert, A.
\newblock {HMDB} 3.0 -- {T}he human metabolome database in 2013.
\newblock \emph{Nucleic Acids Research}, 41\penalty0 (D):\penalty0 801--807,
  2013.
\newblock
  \htmladdnormallink{PMID:23161693}{http://www.ncbi.nlm.nih.gov/pubmed/23161693}.

\bibitem[Yu et~al.(2012)Yu, Chen, Xu, Li, Zhao, Fang, Li, Zhou, Wang, and
  Wang]{Yu12}
Yu, H., Chen, J., Xu, X., Li, Y., Zhao, H., Fang, Y., Li, X., Zhou, W., Wang,
  W., and Wang, Y.
\newblock A systematic prediction of multiple drug-target interactions from
  chemical, genomic, and pharmacological data.
\newblock \emph{PLoS ONE}, 7\penalty0 (5):\penalty0 e37608, 05 2012.
\newblock \doi{10.1371/journal.pone.0037608}.
\newblock URL \url{http://dx.doi.org/10.1371%2Fjournal.pone.0037608}.

\end{thebibliography}
\bibliographystyle{icml2012}
\normalsize

\end{document}